\documentclass[runningheads]{llncs}
\usepackage[T1]{fontenc}
\usepackage{graphicx}
\usepackage{booktabs}
\usepackage[misc]{ifsym}

\usepackage{mwe}
\usepackage{amsmath}
\usepackage{amssymb} 
\usepackage{hyperref} 
\usepackage{pifont} 

\begin{document}

\title{Sugar-Beet Stress Detection using Satellite Image Time Series}
\tocauthor{Bhumika Laxman Sadbhave, Philipp Vaeth, Denise Dejon, Gunther Schorcht, Magda Gregorová}
\toctitle{Loss Functions in Diffusion Models: A Comparative Study}

\titlerunning{Sugar-Beet Stress Detection using SITS}

\author{Bhumika Laxman Sadbhave (\Letter)\orcidID{0009-0004-1129-5729} \and \\
Philipp Vaeth\orcidID{0000-0002-8247-7907} \and \\
Denise Dejon \and 
Gunther Schorcht \and \\
Magda Gregorová\orcidID{0000-0002-1285-8130}}

\authorrunning{B.L. Sadbhave et al.}

\institute{Center for Artificial Intelligence and Robotics, Technical University of Applied Sciences Würzburg-Schweinfurt, Franz-Horn-Straße 2, Würzburg, Germany\\
\email{sadbhavebhumika21@gmail.com, philipp.vaeth@thws.de, dejon@greenspin.de, schorcht@greenspin.de, magda.gregorova@thws.de}}

\maketitle              

\begin{abstract}
Satellite Image Time Series (SITS) data has proven effective for agricultural tasks due to its rich spectral and temporal nature. 
In this study, we tackle the task of stress detection in sugar-beet fields using a fully unsupervised approach. 
We propose a 3D convolutional autoencoder model to extract meaningful features from Sentinel-2 image sequences, combined with acquisition-date-specific temporal encodings to better capture the growth dynamics of sugar-beets. 
The learned representations are used in a downstream clustering task to separate stressed from healthy fields. 
The resulting stress detection system can be directly applied to data from different years, offering a practical and accessible tool for stress detection in sugar-beets.

\end{abstract}

\section{Introduction}

Sugar beet (Beta vulgaris) is a major industrial crop in Europe, particularly in countries like Germany, France, and Poland. 
The European Union accounts for nearly 50\% of the world's sugar-beet production, with Germany playing a significant role in both cultivation and processing~\cite{sugareu}. 
Within Germany, the region of Bavaria is especially important due to its favourable soil, climate, and established farming practices. 
Sugar beet is not only crucial for granulated sugar production but also supports livestock feed, fermentation industries, and biofuel applications, making it economically and environmentally valuable.

Sugar-beet cultivation is subject to a wide range of stress factors including biotic stress such as diseases and pests, and abiotic stress like drought, heat, and nutrient deficiencies. 
These conditions negatively affect both yield and sugar content, and reduce the overall quality and productivity of the crop. 
Therefore, identifying stress within sugar beet fields is essential for gaining insights into crop health and guiding better management practices.

Sugar-beet fields are large and widely distributed, making manual inspection for stress time-consuming and labor-intensive. 
High-resolution imagery and labeled datasets are often costly and impractical for small-scale farmers. 
This study proposes a method for sugar-beet stress detection using publicly available Sentinel-2 satellite data.
The main challenge in this study is the limited availability of labeled data, with only 5\% of sugar-beet fields being labeled. 
Therefore, we use labeled data exclusively for evaluation, and develop a fully unsupervised framework leveraging both spectral and temporal characteristics of Sentinel-2 data.
The code used in this study is available at: \url{https://github.com/bhumikasadbhave/Sugar-beet-Stress-Detection-System.git}.

\section{Related Work}

Traditionally, classical computer vision and machine learning methods have been used for plant disease detection from satellite images, relying on hand-crafted features such as statistical descriptors and vegetation indices like Normalised Difference Vegetation Index (NDVI)\cite{shanmugam2017disease,raza2020exploring,rumpf2010early}. 
Inspired by these approaches, we include histogram-based features as a baseline for comparison with the proposed model. 

Recent studies\cite{victor2024systematic,yu2021early,temp_cnns} have explored deep learning for agricultural tasks like crop classification and disease detection. 
For instance, TempCNNs~\cite{temp_cnns} use 3D convolutions to model temporal satellite data.
These studies treat disease detection as a classification task requiring labeled datasets.
While effective, such supervised methods can be impractical for small-scale industries that lack the resources to label massive datasets.
To address this, recent work\cite{3D_cnns,masked_autoencoder,sat_mae} has applied autoencoders for unsupervised feature extraction from satellite data. 
Additionally, Vision Transformer (ViT) based models\cite{vit_sits,masked_autoencoder,sat_mae} demonstrate the value of temporal encodings in learning data representations. 
Inspired by these studies, we propose an autoencoder model with 3D convolution and temporal encodings derived from acquisition dates for feature learning, followed by clustering for stress detection-- thereby eliminating the need for labeled data.

\subsection{Auto-encoders}
Autoencoders are neural networks used for unsupervised representation learning, enabling tasks such as dimensionality reduction and feature extraction.
They encode the high-dimensional input $\mathbf{x}$ into a lower-dimensional latent representation $\mathbf{z}$ through an encoder,  capturing essential features while discarding irrelevant details. 
A decoder then reconstructs the original input from $\mathbf{z}$, producing $\hat{\mathbf{x}}$.
This process is summarised as:
\[\hat{\mathbf{x}} = \text{Decoder}(\text{Encoder}(\mathbf{x}))\]

\textbf{Loss Function:} The model is trained to minimise reconstruction error between $\mathbf{x}$ and $\hat{\mathbf{x}}$, most commonly using Mean Squared Error (MSE):
\begin{equation}
\mathcal{L}(\mathbf{x}, \hat{\mathbf{x}}) = \frac{1}{n} \sum_{i=1}^{n} \left\| \mathbf{x}_i - \hat{\mathbf{x}}_i \right\|^2 \enspace,
\label{eq:mse}
\end{equation}
where $\mathbf{x}$ is the input data, $\hat{\mathbf{x}}$ is the reconstructed data, and $n$ is the number of examples.
This optimisation enables the model to learn compact, informative representations of the data.

\section{Data}

\begin{figure}[t]
    \centering
    \includegraphics[scale=0.45]{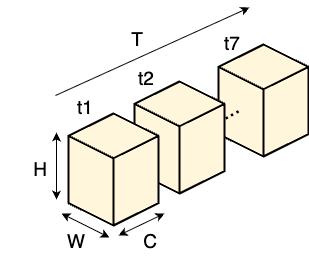}
    \caption{Image data represented as a sequence of temporal instances from $t1$ to $t7$. Each cube corresponds to a single instance with dimensions $(H, W, C)$.}
    \label{fig:data-repr}
\end{figure}

This study uses Sentinel-2 Level 2A images~\cite{sentinel_2a} of agricultural areas in Germany, each containing multiple sugar-beet fields. 
Images have spatial dimensions of $(1000, 1000)$, with float32 reflectance values typically between 0 and 1, occasionally exceeding 1 in highly reflective areas. 
The data covers June to early September 2019, covering the sugar-beet growth cycle from early development to harvest.
We use 10 spectral bands (excluding Sentinel-2 bands 1, 9, and 10 due to redundancy and cloud mask availability), all resampled to 10 m resolution. 
Each image includes multiple temporal instances, from which we select seven cloud-free temporal instances spread accross the growth season.
An image is represented as $I \in \mathbb{R}^{H \times W \times C \times T}$, where $H=1000$, $W=1000$, $C=10$, and $T=7$ denote the height, width, number of channels, and timepoints, respectively (Fig.~\ref{fig:data-repr}).
Three auxiliary masks are used in pre-processing: a cloud mask derived from the Sentinel-2 Scene Classification Map~\cite{sentinel_sceneclassification} to remove clouds, a sugar-beet field ID mask using the field numbers assigned by us for tracking and identification, and an acquisition date mask for temporal encoding. 

The evaluation set for the 2019 season currently includes 35 stressed and 26 healthy sugar-beet fields. 
However, class imbalance remains a potential concern—both in the unlabeled 2019 training set and in future datasets such as the upcoming 2024 season. 
This imbalance may arise from two primary factors. 
First, environmental conditions can lead to widespread stress across fields, making healthy cases relatively rare. 
Second, ground-truth labels are primarily collected by farmers, who tend to report stressed fields more frequently due to their agronomic relevance. 
As a result, the datasets may potentially over-represent stressed cases.

\subsection{Data Preprocessing}

\begin{figure}[t]
    \centering
    \includegraphics[scale=0.36]{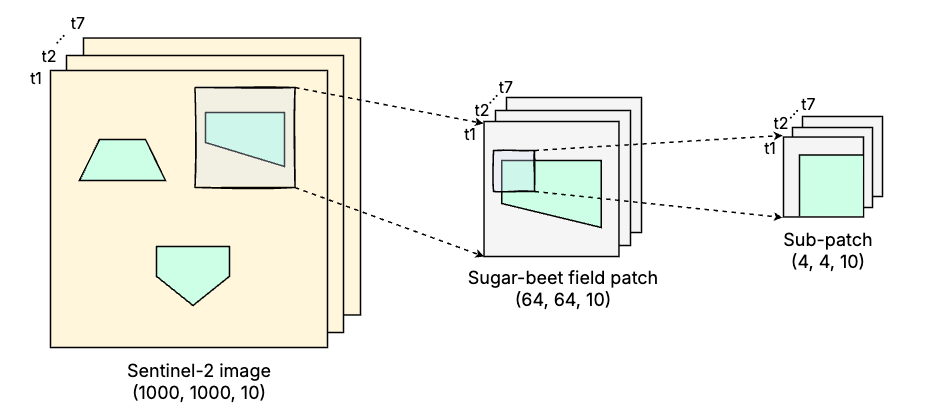}
    \caption{Data preprocessing overview. Each temporal instance is represented as $(H, W, C)$, and the sequence of seven instances is denoted by $t_1, t_2, \dots, t_7$.}
    \label{fig:pre-processing}
\end{figure}

The preprocessing pipeline converts raw Sentinel-2 images into model-ready tensors (Fig.~\ref{fig:pre-processing}). 
We first binarise the sugar-beet field ID mask and apply it pixel-wise across all channels and timepoints to retain only the sugar-beet pixels. 
We then extract each sugar-beet field as a temporal patch and pad it symmetrically with zeros to a uniform spatial size of $(64, 64)$. 
Using the cloud mask, we remove cloud-covered temporal instances, as accurate reflectance values are critical for stress detection. 
We select seven cloud-free instances-- two per month from June to August and one from September-- to ensure seasonal coverage. 
To avoid noise from adjacent vegetation, we exclude border pixels of the sugar-beet fields.

\begin{table}[t]
    \caption{Summary of tensor variants used. Each tensor represents a pre-processed sugar-beet sub-patch over seven temporal instances.}
    \label{tab:tensor_variants}
    \begin{tabular}{lp{9cm}c}
    \toprule
    \textbf{Variant} & \textbf{Channel Composition} & \textbf{Channels} \\
    \midrule
    B10 & All original Sentinel-2 bands (excluding bands 1, 9, 10) & 10 \\
    MVI & NDVI, EVI, and MSI vegetation indices & 3 \\
    B4  & Sentinel Bands 2, 4, 8, 11 (used in NDVI, EVI and MSI) & 4 \\
    \bottomrule
    \end{tabular}
\end{table}
 
Next, we split each patch into non-overlapping $(4, 4)$ sub-patches. 
We discard the empty sub-patches that contain no sugar-beet pixels, and fill the partially empty sub-patches with the channel-wise average of the sugar-beet field pixels within that sub-patch.
We then create three tensor variants using different channel combinations for experimentation, as detailed in Table~\ref{tab:tensor_variants}. 
The vegetation indices Normalised Difference Vegetation Index (NDVI), Enhanced Vegetation Index (EVI), and Moisture Stress Index (MSI) are used as channels in the MVI tensor.
The B10 variant uses all Sentinel-2 bands (except bands 1, 9 and 10), and the B4 variant uses Sentinel-2 bands 2, 4, 8, and 11-- used to calculate vegetation indices as channels.
Each sub-patch is represented as a 4D tensor \( \mathbf{x} \in \mathbb{R}^{4 \times 4 \times C \times 7} \), where C depends on the tensor variant. 

\section{Model}

Temporal information is essential for understanding the sugar-beet growth and evolution of stress patterns. 
Although Sentinel-2 provides frequent images, cloud cover often limits its usability. 
We select seven non-clouded images per field spread accross the growth season. However, the acquisition dates vary across sugar-beet fields and the temporal sequences are not uniformly spaced.
We address this by adding acquisition-date-specific temporal encodings to the input tensor. 
We use sinusoidal encodings~\cite{vaswani2017attention} to represent acquisition dates, mapping each day $d \in [1, 365]$ to a continuous annual cycle. 
The encodings are computed as follows:

\[\text{e}_{s} = \sin\left( \frac{2\pi d}{365} \right), \quad \text{e}_{c} = \cos\left( \frac{2\pi d}{365} \right)\]

These are added element-wise to all pixels of the input data tensors during the forward pass.

The modeling process comprises of feature extraction, downstream clustering, and converting sub-patch predictions to field-level labels. 

\begin{figure}[t]
    \centering
    \includegraphics[scale=0.55]{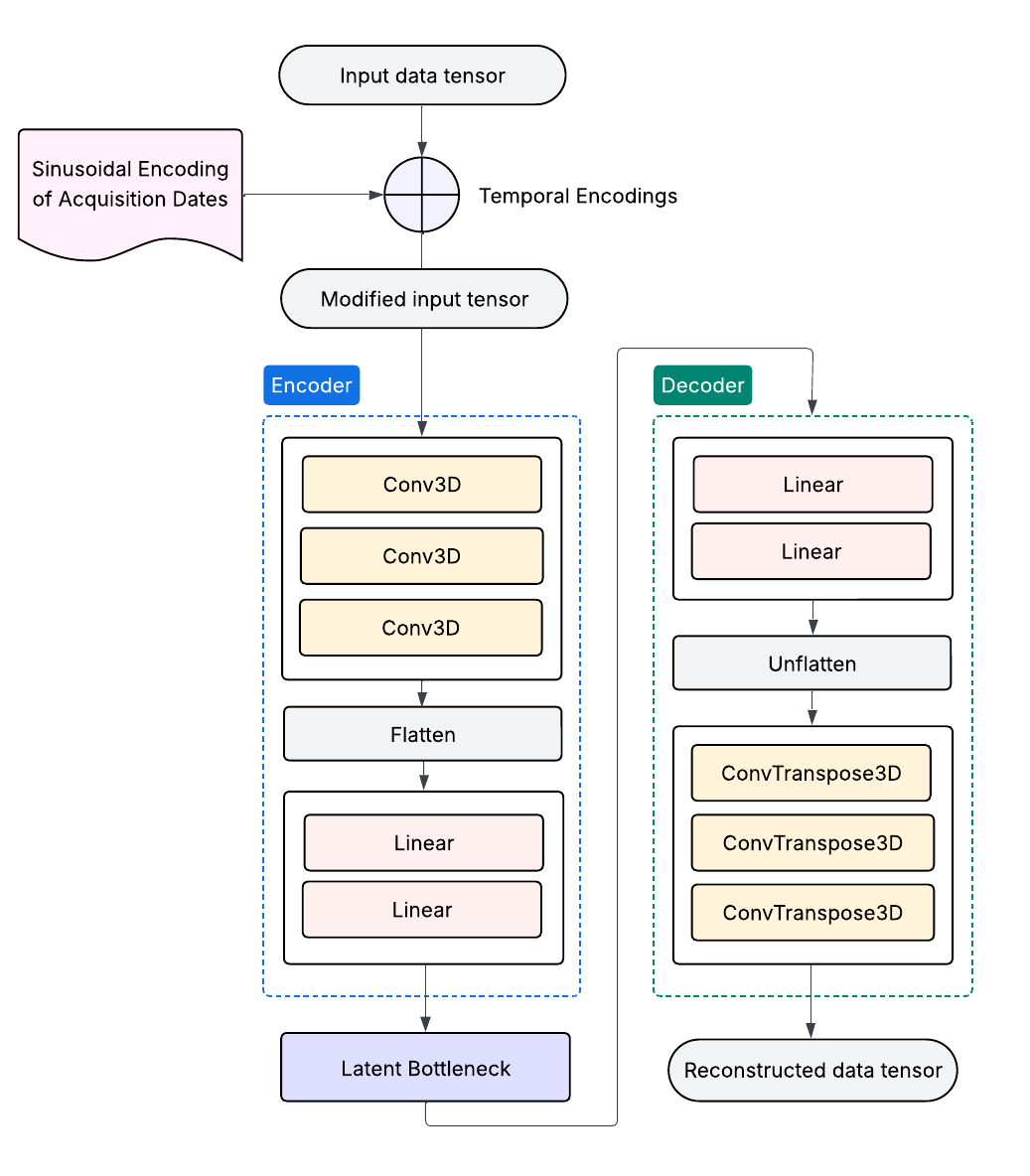}
    \caption{3D\_AE\_B10 model with temporal encodings.} 
    \label{fig:architecture}
\end{figure}

To obtain compact and meaningful representations from the preprocessed sub-patch tensors, we use the autoencoder architecture, with 3D convolutions. 
Temporal encodings are first added to the input tensor, which is then passed through 3D convolutional layers that preserve the spatiotemporal dimensions while progressively increasing the number of feature maps, to capture both temporal and spectral characteristics. 
The resulting tensor is flattened and passed through fully connected layers to produce a latent representation $\mathbf{z} \in \mathbb{R}$.
The decoder reconstructs the input tensor from the latent representation using transposed 3D convolutions, thus restoring the original dimensions. 
We train the model using the MSE loss (equation~\eqref{eq:mse}), after which, we extract the latent representations and use them as feature vectors for downstream clustering.
We use B10 tensor as the default and refer to this model as 3D\_AE\_B10 (Fig.~\ref{fig:architecture}).

Due to limited labeled data, we categorise the sugar-beet sub-patches as stressed or healthy by applying unsupervised clustering on the learned features \(\mathbf{z}\) instead of supervised classification.
We use k-means to group the features into two clusters, which we interpret as the target classes.

Since ground truth labels are only available for the entire sugar-beet fields, we aggregate sub-patch predictions using a threshold \( \alpha \in [0, 1] \).  
If more than \( \alpha\% \) of the sub-patches are predicted as stressed, the entire sugar-beet field is labeled stressed.  
We use \( \alpha = 0.5 \) as the default value and analyse the effect of varying this threshold in the experiment section.

\section{Experiments}\label{section:experiments}
A total of 1857 unlabeled cloud-free sugar-beet fields corresponding to 1228 patches are partitioned into 33128 sub-patches. 
Of these, 80\% are used for training and 20\% for testing the 3D\_AE\_B10 model. 
The latent features extracted from all 1857 unlabeled fields are then used for k-means clustering. 
The evaluation set comprises of 61 cloud-free sugar-beet fields corresponding to 48 patches, further divided into 1197 sub-patches.

Model performance is calculated over sugar-beet fields, specifically across the 61 fields in the evaluation set, using both accuracy and F1-score.
While accuracy reflects the overall correctness, the F1 score provides a deeper insight, since the primary focus is detecting stressed fields while minimizing false positives.
Due to potential class imbalance in the dataset, as highlighted in the data section, accuracy by itself can be misleading.
The results are averaged over three independent executions to account for randomness in initialisation and training.

\begin{table}[t]
    \caption{K-means clustering results on sugar-beet fields using various feature extraction methods.}
    \label{tab:model_comparison}
    \begin{tabular}{p{4cm} c c c c c}
    \toprule
    \textbf{Method} & \textbf{Data} & \textbf{Temporal} & \textbf{Threshold} & \textbf{Accuracy} & \textbf{F1-score} \\
     & \textbf{Tensor} & \textbf{Encodings} & \textbf{$\alpha$} & \textbf{(\%)} & \textbf{(\%)} \\
    \midrule
    Raw data & B10 & \ding{55} & 0.5  & 63.93 & 74.42 \\
    Histogram features & B10 & \ding{55} & 0.5 & 63.93  & 69.7 \\
    2D\_AE\_B10 & B10 & \ding{55} & 0.5 & 60.11 & 67.78 \\
    3D\_AE\_B10 & B10 & \ding{55} & 0.5 & 59.56 & 67.25 \\
    3D\_AE\_B10 & B10 & \ding{51} & 0.5 & \textbf{69.40} & \textbf{75.21} \\
    \midrule
    3D\_AE\_MVI & MVI & \ding{51} & 0.5 & 52.46 & 64.32 \\
    3D\_AE\_B4 & B4 & \ding{51} & 0.5 & 60.65 & 68.84 \\
    \bottomrule
    \end{tabular}
\end{table}

\subsection{Modeling Results}

The proposed 3D\_AE\_B10 model is evaluated against three baselines. 
The first applies k-means directly on the sub-patch tensors, without any feature abstraction. 
The second uses k-means on histogram features~\cite{histogram} computed across all temporal instances. 
The third, 2D\_AE\_B10, is a 2D convolutional variant of the 3D\_AE\_B10 model.
It processes input with temporal instances stacked along the channel dimension and omits temporal encodings, resulting in shape $\mathbb{R}^{H \times W \times C \cdot T}$. 
We also evaluate the 3D\_AE\_B10 model without temporal encodings to assess their impact on performance.

As shown in Table~\ref{tab:model_comparison}, 3D\_AE\_B10 with temporal encodings achieves the highest F1-score, indicating that both 3D convolutions and temporal encodings help to capture meaningful spatiotemporal patterns.
It is however closely followed by the raw data representation, indicating that the raw input may already contain strong discriminative signals. 
Removing temporal encodings leads to a clear drop in performance for the 3D variant.
The 2D\_AE\_B10 model also performs poorly, with an F1-score similar to the 3D variant without encodings. This suggests that treating temporal frames as channels is not sufficient to capture temporal dynamics.
The histogram-based approach scores lower, likely due to loss of detail during feature aggregation.
Overall, F1-score comparisons show that temporal encodings and 3D structure lead to useful, although modest improvements in performance, highlighting the value of explicitly incorporating the temporal information.

While B10 is the primary input for model comparison, we additionally evaluate the 3D\_AE architecture using two alternative tensors: MVI with multiple vegetation indices, and B4 with four Sentinel-2 bands used to compute MVI (Table~\ref{tab:tensor_variants}). 
The result of this experiment suggests that the choice of input tensor impacts the model performance significantly.
As shown in Table~\ref{tab:model_comparison}, the B10 tensor achieves the highest F1-score, highlighting the value of its richer spectral content for feature learning.
In comparison, the MVI tensor performs worst, showing that vegetation indices alone may lack the spectral detail needed for effective unsupervised learning. 
Interestingly, the B4 tensor outperforms MVI, indicating that even limited raw band data can retain more useful information than pre-computed vegetation indices.

\subsection{Effect of varying the sub-patch-to-patch threshold $\alpha$}
\begin{figure}[t]
    \centering
    \includegraphics[scale=0.34]{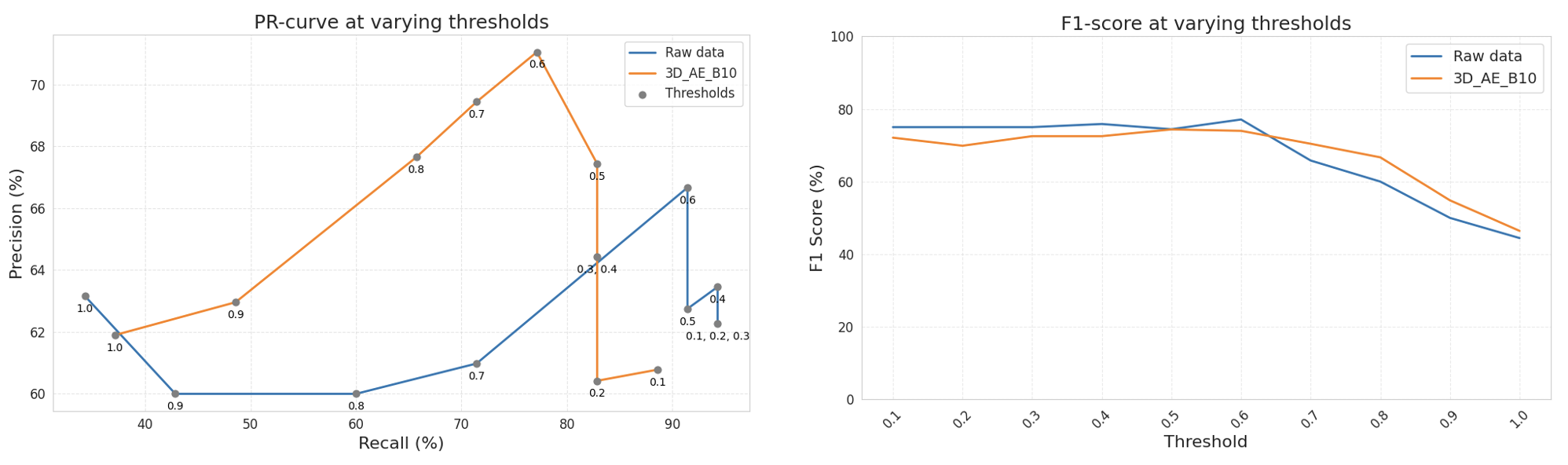}
    \caption{Precision-Recall curve for varying the sub-patch-to-patch threshold $\alpha$.}
    \label{fig:threshold_curve}
\end{figure}

We further examine the impact of varying the sub-patch-to-patch threshold $\alpha \in [0,1]$. 
A sugar-beet field is labeled as stressed if the proportion of its sub-patches predicted as stressed exceeds the threshold $\alpha$.
Fig.~\ref{fig:threshold_curve} presents the precision-recall and F1-score curves for the 3D\_AE\_B10 model and the best-performing baseline (raw data), with $\alpha$ varying from $0.1$ to $1.0$. 
 
At lower thresholds, the baseline achieves higher recall than 3D\_AE\_B10 and nearly equivalent F1-scores, highlighting its strength in detecting stressed fields. 
As the threshold increases, a larger proportion of sub-patches must be predicted as stressed for the entire field to be labeled as stressed. 
Under these stricter conditions, 3D\_AE\_B10 begins to outperform the baseline, particularly in precision—and consequently in F1-score. 
This shift reflects the tendency of the baseline model to incorrectly label healthy fields as stressed when the threshold is high. 
While both models perform comparably around the default threshold $\alpha=0.5$, 3D\_AE\_B10 shows greater robustness at higher thresholds, maintaining better precision and a slightly higher F1-score. 
This suggests that the spatiotemporal representations learned by 3D\_AE\_B10 lead to more reliable stress detection under higher thresholds.

\section{Conclusion}
In this study, we developed a fully unsupervised system for stress detection in sugar-beet fields using publicly available satellite images. 
The approach requires no labeled training data and specifically highlights the importance of incorporating temporal acquisition dates as part of the model input. 
Given the limited availability of labeled data, the current results are exploratory in nature. Nevertheless, they demonstrate the potential of unsupervised methods to generate actionable insights in scenarios with limited labeled data. 
Moreover, the developed stress detection system is designed to generalise across years, making it easily deployable on future datasets.

\subsubsection{Future Work:}
Once labeled data becomes available, we plan to transition from downstream clustering to supervised classification, enabling more reliable model performance and clearer benchmarking.
We also aim to validate our system on the 2024 sugar-beet season once the corresponding ground truth labels are available. 
This study forms part of a broader effort focused on early stress detection in sugar-beet fields, with the goal of enabling timely interventions and reducing crop losses. 
As part of this effort, we will explore using temporal data from June to July instead of the full sugar-beet growth cycle from June to September, for facilitating earlier detection.
Additionally, while the current pipeline operates at the sub-patch level, future work will explore pixel-level stress detection for more fine-grained analysis.

By providing a reusable and generalisable framework, this study lays the groundwork for a more robust stress detection system, and for advancing unsupervised representation learning in agriculture.



%
%
%
\bibliographystyle{splncs04}
\bibliography{references}
%




\end{document}